\documentclass[a4paper]{article}
\usepackage{setspace}
\usepackage[ruled,linesnumbered,noline]{algorithm2e}
\usepackage{amssymb,amsmath,epsfig,latexsym,graphicx,bm}

\usepackage[left = .75in, top=.75in,right=.75in,nohead,nofoot]{geometry}
\usepackage{epstopdf}
\include{newcommand}
\title{GreMuTRRR: A Novel Genetic Algorithm to Solve Distance Geometry Problem for Protein Structures}

\author{ Md. Lisul Islam$^{1,2}$, Swakkhar Shatabda$^1$ and M Sohel Rahman$^2$\\
$^1$Department of Computer Science and Engineering, United International University\\
House 80, Road 8A, Dhanmondi, Dhaka-1209, Bangladesh\\ $^2$Department of Computer Science and Engineering, \\Bangladesh University of Engineering and Technology, Palashi, Dhaka-1000, Bangladesh\\
Email: \{lisul,swakkhar\}@cse.uiu.ac.bd, 
msrahman@cse.buet.ac.bd
}

\date{}

\begin{document}

\maketitle
\doublespacing

\begin{abstract}
Nuclear Magnetic Resonance (NMR) Spectroscopy is a widely used technique to predict the native structure of proteins. However, NMR machines are only able to report approximate and partial distances between pair of atoms. To build the protein structure one has to solve the Euclidean distance geometry problem given the incomplete interval distance data produced by NMR machines. In this paper, we propose a new genetic algorithm for solving the Euclidean distance geometry problem for protein structure prediction given sparse NMR data. Our genetic algorithm uses a greedy mutation operator to intensify the search, a twin removal technique for diversification in the population and a random restart method to recover stagnation. On a standard set of benchmark dataset, our algorithm significantly outperforms standard genetic algorithms.
 
\end{abstract}

\vspace{6mm}
{\bf keywords-}
Genetic Algorithms, Protein Structures, NMR, Euclidean Distance Geometry, Greedy Mutation

\section{Introduction}
The function of a protein depends on its \textit{native conformation} or the stable three dimensional structure with minimum free energy in a particular environment. Knowledge about this native structure is of paramount importance and can have an enormous impact on the field of rational drug discovery. Nuclear magnetic resonance (NMR) spectroscopy is one of the most widely used techniques to predict native structure of proteins. NMR machines provide inter-atomic distance data for a given protein. Thus, the molecular distance geometry problem (MDGP) arises in this context: given the set of Euclidean distances between the atoms in a protein, MDGP asks to find the Cartesian coordinates of the atoms.  

However, in practice, NMR machines are able to produce the inter-atomic distances of only a subset of the pairs of atoms that are spatially close and this data too lacks accuracy. As a result, we are given the upper and lower bounds of only a subset of the Euclidean distances. Thus we are left to solve a variant of the MDGP problem with incomplete and inaccurate data. Many computational approaches have been applied to solve MDGP problem with sparse and inaccurate data on real instances \cite{souza2013solving,liberti2014euclidean,wu2007updated,mucherino2009comparisons,liberti2011molecular,more1999distance}. However, complete search methods like spatial branch and bound (sBB) and stochastic methods like variable neighborhood search (VNS) have been able to solve the problem for proteins with only upto 50 atoms \cite{lavor2006computational} and fail to converge quickly for larger number of atoms. 

In this paper, we present GreMuTRRR (pronounced grey matter), which is a scalable genetic algorithm to solve the MDGP problem for sparse and inaccurate NMR data. Much of the success of our approach comes from a greedy mutation operator, sude to intensify the search, a twin removal technique to diversify the population and a random restart method to recover from stagnation. Experimental results on proteins with number of atoms ranging from 50--2147 shows that our algorithm outperforms standard genetic algorithms and thus obtains state-of-the-art results.

\section{Preliminaries}
In this section, we provide a formal definition of the MDGP problem and a brief description of genetic algorithms.  
\subsection{Distance Geometry Problem}
Molecular distance geometry problem can formally be defined as followed \cite{crippen1988distance}: find a set of Cartesian coordinates $c_1, c_2, \cdots, c_n \in \mathbb{R}^3$ of atoms of a molecule such that $l_{ij} \le d_{ij} \le u_{ij},\forall(i,j)\in E$, where $l_{ij}$ and $u_{ij}$ are the lower and upper bounds of the Euclidean distance $d_{ij}\equiv ||c_{i}-c_j||$ between a pair of atoms $(i,j)\in E$. Notably, in the context of NMR data, $E$ is sparse. Here, Cartesian coordinate $c_i$ of an atom $i$ corresponds to a three dimensional point $(x_i,y_i,z_i)$ in the Euclidean space. 
\subsection{Genetic Algorithm}
Genetic Algorithms are population-based search methods that resemble the natural phenomena of biological evolution. Genetic Algorithms are widely used for different search optimization problems in different fraternity. It basically starts with a pool of initial random solutions which is called the initial population. Each individual in the population are encoded by a set of properties which are called chromosome or genotype which can be altered for attaining diversification in the population. It then follows an iterative process in which each of these iteration is called a generation. In each generation, the population are then allowed to evolve using different operators which also mimics the natural process of biological evolution like mutation, recombination or survival of the fittest. In each generation, the fitness of each individual is evaluated. Generally the fitness is the value of the optimization function being considered to be solved. The more fit individuals are usually selected to breed and generate new fitter individuals in the population. Thus a new generation of population is `breeded' and are used in the next generation. This process of evolution are continues until a sufficient number of generations has been produced or a satisfactory level of fitness value has been attained.  
\section{Related Works}
Euclidean distance geometry problem and its variants are applied to many problems in various fields including wireless sensor network localization \cite{savarese2001location}, inverse kinematic problem \cite{tolani2000real}, multi dimensional scaling \cite{tenenbaum2000global}, protein structure determination \cite{more1999distance} etc. The variant of the MDGP problem, when we know distances for all pairs $(i,j)\in E=\{1,2,\cdots\}^2$ and  $d_{ij}=l_{ij}=u_{ij}$ has a polynomial algorithm to produce exact solution \cite{crippen1988distance}. Even when some of the pairwise distances are unknown, the problem is solvable by a linear time algorithm \cite{wu2007updated}. However, the variant of MDGP with sparse and inaccurate data is shown to be NP-hard by More and Wu \cite{more1997global}. A recent survey of computational methods applied to solve this variant of MDGP can be found in~\cite{liberti2014euclidean}. 

Among the general purpose methods, spatial branch and bound \cite{liberti2005comparison} and variable neighborhood search (VNS) \cite{liberti2005variable} methods are not scalable \cite{lavor2006computational}. Smoothing based methods like DGSOL~\cite{more1999distance,more1997global} also fail for large instances of the problem. In \cite{liberti2009double}, VNS was combined with DGSOL approaches which provided better results for larger instances but resulted into a slow algorithm. A combinatorial build up algorithm was proposed in \cite{dong2003geometric}. It is important to note that all these methods were tested on dense instances only. Among other methods applied to this problem graph decomposition methods \cite{souza2013solving} and NLP formulations \cite{hendrickson1995molecule} are notable.

\section{Our Method}
GreMuTRRR is formally presented in Algorithm~\ref{algoMain}. It starts with a population initialized randomly and terminates at convergence. Convergence is achieved if the search does not improve the quality of the global solutions for a given period of time. In each generation, individuals are selected from the population for cross-over to produce new individuals. Mutation operations are performed probabilistically on the newly found individuals. Global best solutions are updated in each iteration and are kept in the new populations to maintain elitism. Periodical twin removal and random restart operations are activated to ensure diversification and recover stagnation. Rest of this section describes the various components of GreMuTRRR.

\begin{algorithm}
\begin{spacing}{0.8}
\DontPrintSemicolon
$nonDiverseSteps=0$\;
$nonImprovingSteps=0$\;
$GreedyMutationRate = 0.8$\;
Intialize the population, $P$ randomly\;
\While{Termination criterion are met}
{
	$P_{new}=\{globalBest\}$\;
	
	\For{each Individual $X \in P$}
	{
		$\langle X_1,X_2\rangle = \mathsf{tournamentSelection}(P$)\;
		$X_{new} = \mathsf{crossOver}(X_1,X_2$)\;
		add $X_{new}$ to $P_{new}$\;
	}
	\For{each Individual $X$ in $P_{new}$}
	{
		\eIf{$rand(0,1) \leq GreedyMutationRate$}
		{
			$\mathsf{greedyMutate}(X)$\;
		}
		{
			$\mathsf{randomMutate}(X)$\;
		}
	}
	find the individual $X_{best}\in P_{new}$ with best fitness \;
	\eIf{$\mathsf{fitness}(globalBest) < \mathsf{fitness}(X_{best}) $}
	{
			{
			$globalBest=X_{best}$\;
			$nonImprovingSteps=0$}\;
	}
	{
		$nonImprovingSteps++$\;
	}
	\eIf{ $nonDiverseSteps \ge twinRemovalInterval$ }
	{
		activate $\mathsf{twinRemoval}(P_{new})$ procedure\;
		$nonDiverseSteps=0$
	}
	{
		$nonDiverseSteps++$\;
	}
	\If{$nonImprovingSteps \ge randomRestartInterval $}
	{
		activate $\mathsf{randomRestart}(P_{new})$ procedure\;
		$nonImprovingSteps=0$
	}
	$P=P_{new}$\;
}
	\Return $globalBest$\;
\caption{GreMuTRRR()\label{algoMain}}
\end{spacing}
\end{algorithm}

\subsection{Search Space}	
A Protein structure is formed by the interaction among different amino acids tied with each other by chemical bonds. {A pair of these amino acid cannot reside closer than 3.8{\AA} to each other to avoid steric clash.} If all the amino acids in a certain protein structure align in a straight line, we have a chain of amino acids of length $3.8 \times V$ in each directions, where $V$ is the number of amino acids in the protein structure. Hence, we have taken the upper bound of our 3-D search space to $3.8 \times V$.
\subsection{Encoding}
In our method, we have encoded every individual in a population by \textit{$3 \times V$} number of genes. Here, $V$ is the number of atoms in a protein structure. Each of these genes in an individual is initialized randomly from an uniform distribution of the range $[0,3.8 \times V]$. So, an individual $X$ can be represented as the ordered list of \textit{$3V$} number of genes:
\begin{footnotesize}
	\[
	X = \left\{ \underbrace{x_1,y_1,z_1}_{1^{st} Atom},\underbrace{x_2,y_2,z_2}_{2^{nd} Atom},\ldots,
	\underbrace{x_{i},y_{i},z_{i}}_{i^{th} Atom},\ldots,\underbrace{x_{V},y_{V},z_{V}}_{V^{th} Atom}\right\}
	\]
\end{footnotesize}

\subsection{Fitness Evaluation}
We have calculated the euclidean distance between each pair of points present in the individual's chromosome where each non-overlapping consecutive subsequence of length three in the chromosome represents the position of an atom in the protein structure in a 3-D space. We have assumed an upper bound, $ u_{ij}$ and lower bound, $ l_{ij}$ of the distances between pair of amino acids. 
The fitness of an individual $X$ is defined by as Equation~\ref{equ1} below:

\begin{equation}
\begin{scriptsize}
	Fitness(X) = {\left( \frac{1}{\vert E \vert} \displaystyle\sum_{(i,j)\in E} e_{ij}^{2} \right)}^{1/2}
\label{equ1}
\end{scriptsize}
\end{equation}

	where, 
	\begin{footnotesize}
	$	e_{ij} = max\lbrace{l_{ij}-\Vert{c_i-c_j}\Vert,\Vert{c_i-c_j}\Vert - u_{ij},0}\rbrace
	$
	\end{footnotesize}
	is the error associated to the constraints $l_{ij}\leq \Vert{c_i-c_j}\Vert \leq u_{ij}$ and ${\vert E \vert}$ denotes the number of distance pairs given. This is known as the \textit{Largest Distance Error (LDE)} in the literature~\cite{liberti2014euclidean}. 
\subsection{Genetic Operators}
Genetic algorithms are guided by different genetic operators to help in keeping a balance of the exploitation and exploration of the search process. We use three types of genetic operators in our search: random mutation, greedy mutation and cross-over.        
\subsubsection{ Random Mutation}
Genetic diversity is a necessity for the process of evolution which is generally attained by mutation operator. Mutation helps the evolutionary process to guide through and look for different avenues and solutions in the search space. It also helps to avoid local optima by preventing the individuals from becoming too similar to one another.  We have used uniform mutation in our method. We have mutated the value of each gene into a new value within the pre-specified bound of the search space($[0,3.8 \times V]$). Whether an individual will be mutated or not depends on a mutation rate which is kept to a lower value of 0.015 to avoid primitive random search. A sketch of the pseudo-code is given in Algorithm~\ref{algoRM}.	
	\begin{algorithm}
\begin{spacing}{0.8}
		\DontPrintSemicolon
		$mutationRate = 0.015 $\;
		\For{each gene X(i) in the genotype of $X$}
		{
			\If{$rand(0,1) \leq mutationRate$}
			{
				$ X(i) = U(0,1)\ast (3*V);$
			} 
		}
		\Return $X$
		\caption{Random Mutation (individual $X$)\label{algoRM}}
		\end{spacing}
		\end{algorithm}

\subsubsection{Greedy Mutation}
In our proposed method, we have also used a greedy mutation where we greedily choose the new value of a particular gene in an individual. We randomly alter a certain gene and try $r$ different random values for the gene temporarily. The value of gene that gives the best fitness of that individual is retained and used as the final new value of the gene. As, greedy mutation is computationally expensive, we make selection between the random mutation and greedy mutation with a probability, $GreedyMutationRate(=0.8)$. The algorithm for greedy mutation is given in Algorithm~\ref{algoGM}.
\begin{algorithm}
\begin{spacing}{0.8}
\DontPrintSemicolon
set $r = 20$\;
\For{each gene $X(i)$ in the genotype of $X$}
{
	S = set of $r$ random values for gene $X(i)$ from the range $[0,3.8\ast V]$\;
	find $v \in S$ for which fitness(X) is minimum\;
	set, $X(i) = v$\;
}
\Return $X$
\caption{Greedy Mutation (individual $X$) \label{algoGM}}
\end{spacing}
\end{algorithm}
\subsubsection{Cross-over}
Cross-over operators help genetic process to exploit the better solutions found thus far along the evolution process and regenerate new individuals by combining the genetic information of the individuals with better fitness value. It uses the better historical genetic information to guide the search into the search space regions with solutions having better fitness. By recombining the fitter individuals to generate new offspring, this operator is likely to produce more fitter	individuals. We have used tournament selection with tournament size being equal to 5 to select two individual parents to take part in the cross-over operation. We then have recombined genes of the two parents and have generated a new set of genes for each of the offsprings with equal probability for each gene of being selected from any of the parents. The algorithm is outlined in Algorithm~\ref{algoCO}.
\begin{algorithm}
\begin{spacing}{0.8}
\DontPrintSemicolon
$X_{new}$ be a new offspring\; 
$uniformRate = 0.5 $\;
\For{each gene X(i) in the genotype of $X_{new}$}
{
	\eIf{$rand(0,1) \geq uniformRate$}
	{
		$X_{new}(i) = X_1(i)$\;
	}
	{
		$X_{new}(i) = X_2(i)$\;
	}
}
\Return $X_{new}$\;
\caption{Cross-Over (individual $X_1$, individual $X_2$)\label{algoCO}}
\end{spacing}
\end{algorithm}

\subsection{Twin Removal}
We have removed and reinitialized individuals with identical genetic information. We have defined the similarity measure between two individuals as follows:
\begin{equation}
		Similarity(X_1,X_2) = e^{-\frac{{\Vert X_1-X_2 \Vert}^2}{2\sigma^2}}
\end{equation}

Here, the value of $\sigma$ has been chosen to be 75. The $Similarity(X_1,X_2)$ function will return a value in the range [0,1]. The more the value of the similarity function is closer to 1, the more genetically similar are $X_1$ and $X_2$. We have chosen 0.8 as the threshold to define similarity between two individual. If $Similarity(X_1,X_2) >= 0.8$, we declare that $X_1$ and $X_2$ are twins and  reinitialize randomly one of them. We have run the this Twin Removal procedure after every 100 generations of evolution.
\begin{algorithm}
\begin{spacing}{0.8}
\DontPrintSemicolon
$similarityThreshold = 0.8$\;
\For{each pair of individuals $(X_i,X_j)$ in the population}
{
	\If{$Similarity(X_i,X_j) \geq similarityThreshold$}
	{
		declare $X_i$ and $X_j$ as Twins\;
		reinitialize $X_j$\;
	}
}
\caption{Twin Removal}
\end{spacing}
\end{algorithm}
\subsection{Random Restart}
If the algorithm does not show enough improvement within a significant number of generations, we reinitialize some individuals of the population with random values within the search space. We have evaluated the improvement over the immediate past 50 generations. If the improvement in the last 50 generations is less than or equal to a threshold, $t = 0.001$, we initiate the random restart procedure.   
\section{Experimental Results}
We have implemented GreMuTRRR in Java programming language using JDK 1.6 and have run our experiments on an Intel 3.3GHz core $i$3 machine with 2GB RAM. We compared the performance of GreMuTRRR with a basic genetic algorithm without considering the greedy mutation, twin removal and random restart. We denote this algorithm as `Basic GA' henceforth. Basic GA differs with GreMuTRRR given in Algorithm~\ref{algoMain} in Lines 27--36 since it does not contain twin removal and random restart and in Line 3, as the parameter $greedyMutationRate=0.0$ for Basic GA. All other parameters were kept same for the sake of fair comparison. 
\subsection{Benchmark}
We have considered two sets of protein instances for our experiments: one considering only the backbone atoms (backbone only) and another containing all atoms (full atomic). These are larger protein benchmarks introduced in \cite{biswas2008distributed}. We extracted the structures from PDB website \cite{PDB} and have calculated the distance among pair of atoms. Lower and upper bounds of the distances are calculated using the following equation:
\begin{equation}
	l_{ij} = (1-\epsilon)\hat{d}_{ij}, 
		u_{ij} = (1+\epsilon)\hat{d}_{ij} 
\end{equation}
Here, $\hat{d}_{ij}$ is the real distance between point $c_{i}$ and point $c_{j}$ in the known structure of the protein sequence and $\epsilon=0.8$. To make the dataset sparse, we took only $30\%$ of the distances that are less or equal to 6$\AA$.   
\subsection{Results}
The largest distance error (LDE) values for our algorithm and Basic GA for backbone only instances and full atomic instances are given in Table~\ref{table1} and Table~\ref{table2} respectively. Each protein is reported with its corresponding PDB id and number of atoms in the structure considered.
\begin{table}[ht]
\caption{Best Fitness for population size 50 with run of 2000 generations(with backbone atoms only)}
\centering
\begin{tabular}{c c c c} 
\hline\cline{1-4} 
Protein Id & $V$ & Basic GA & GreMuTRRR \\ [0.5ex] 
\hline 
1PTQ 	& 50   	& 0.49546&	0.44977 \\ 
1LFB 	& 77	& 0.18060&	0.13360 \\
1F39 	& 101 	& 0.06595&	0.02051 \\
1AX8 	& 130 	& 0.06064&	0.00253\\
1RGS 	& 264 	& 0.12380&	0.00297\\
1TOA 	& 277 	& 0.12993&	0.00156\\
1KDH 	& 356 	& 0.19119&	0.00458\\
1BPM 	& 481 	& 0.34434&	0.00441\\
1MQQ 	& 679 	& 0.58863&	0.00894\\ 

\hline 
\end{tabular}
\label{table1} 
\end{table}

\begin{table}[ht]
\caption{Best Fitness for population size 50 with run of 1500 generations(including all atoms)}
\centering
\begin{tabular}{c c c c} 
\hline\cline{1-4} 
Protein Id & $V$ & Basic GA & GreMuTRRR \\ [0.5ex] 
\hline 
1PTQ 	& 402   	& 0.71000	& 0.17224 \\ 
1LFB 	& 641		& 0.97121	& 0.31463\\
1F39 	& 767 		& 1.81091	& 0.16805\\
1AX8 	& 1003 		& 2.52303	& 0.43211\\
\hline 

\end{tabular}
\label{table2} 
\end{table}
\subsection{Analysis}
\begin{figure}[t]
\begin{center}
\includegraphics[scale=0.4]{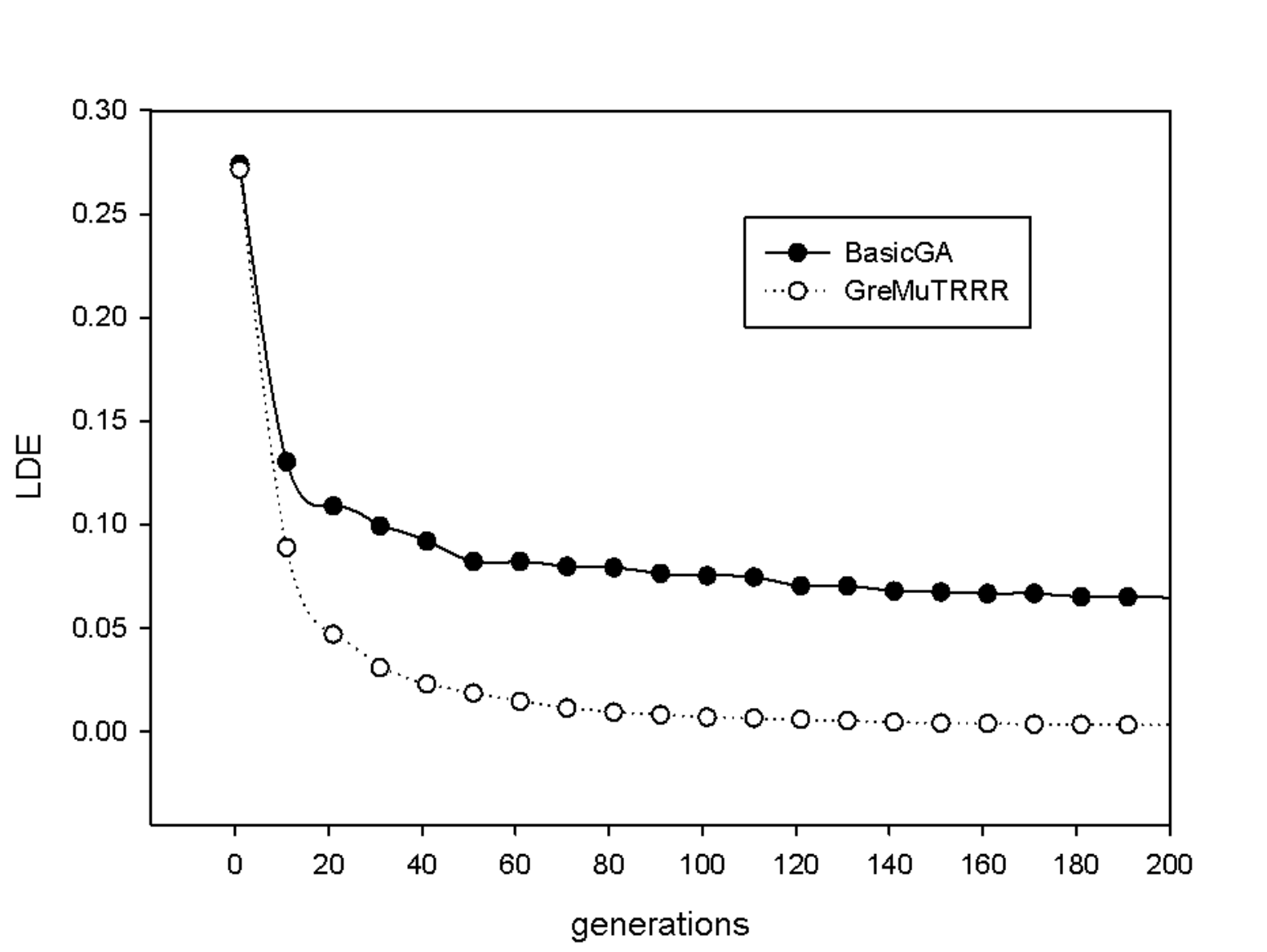}
\end{center}
\caption{ Search progress for two algorithms for the protein 1AX8.\label{figMain}}
\end{figure}

From the results presented in the tables it is evident that for both of the instances GreMuTRRR achieves better solutions compared to the Basic GA algorithm for the proteins. However, the improvement is more clearly shown in Figure~\ref{figMain}. GreMuTRRR significantly lowers the objective function LDE before both algorithms gets stuck and can not improve further. Its interesting to note thatGreMuTRRR is able to converge very early ($\le$ 50 generation) compared to the Basic GA.
\section{Conclusion}
In this paper, we presented a new genetic algorithm to solve the molecular distance geometry for protein structure determination problem using NMR data. We use a greedy mutation operator to intensify the search, a twin removal technique for diversification in the population and a random restart method to recover stagnation. On a standard set of benchmark dataset, our algorithm significantly outperforms standard genetic algorithms. In future, we want to implement a web service based on our method and make it public to be used by biologists.
\begin{spacing}{0.86}
\bibliographystyle{abbrv}
\bibliography{icece2014}
\end{spacing}
\end{document}